\documentclass[conference]{IEEEtran}
\IEEEoverridecommandlockouts
\usepackage[left=1.62cm,right=1.62cm,top=1.9cm]{geometry}

\usepackage{times}
\usepackage{soul}
\usepackage{url}
\usepackage[hidelinks]{hyperref}
\usepackage[utf8]{inputenc}
\usepackage[small]{caption}
\usepackage{graphicx}
\usepackage{amsmath}
\usepackage{amsthm}
\usepackage{booktabs}
\usepackage[switch]{lineno}
\usepackage{url}
\usepackage{mathtools}
\usepackage{amssymb}
\usepackage{soul}

\usepackage{xcolor}
\usepackage{soul}
\usepackage{titlesec}
\usepackage[acronym]{glossaries}
\usepackage{comment}
\usepackage[normalem]{ulem}

\usepackage{caption}

\usepackage{soul}
\usepackage[linesnumbered,ruled]{algorithm2e}
\usepackage{titlesec}
\usepackage[acronym]{glossaries}
\usepackage{comment}
\titlespacing{\section}{0pt}{\parskip}{-\parskip}

\newcommand{\sps}[2]{\hl{#1}\textcolor{magenta}{[#2]}}
\newcommand{\kpr}[2]{\hl{#1}\textcolor{green}{[#2]}}

\newacronym{fl}{FL}{federated learning}
\newacronym{ai}{AI}{artificial intelligence}
\newacronym{ros}{ROS}{robot operating system}
\newacronym{ml}{ML}{machine learning}
\newacronym{pid}{PID}{propotional-integral-derivative}
\newacronym{px}{px}{pixels}
\newacronym{fov}{FoV}{field-of-view}
\newacronym{roi}{RoI}{region-of-interest}
\newacronym{trl}{TRL}{technology readiness level}

\newcommand{\degree}{^{\circ}}

\newcommand{\maze}{M}
\newcommand{\robot}{R}
\newcommand{\server}{S}
\newcommand{\A}{\alpha}
\newcommand{\B}{\beta}

\begin{document}
%\captionsetup[figure]{name={Fig.},labelsep=period}

\title{Maze Discovery using Multiple Robots via Federated Learning
}
\author{
\IEEEauthorblockN{
Kalpana~Ranasinghe\IEEEauthorrefmark{1}, 
H.P.~Madushanka\IEEEauthorrefmark{1},~\IEEEmembership{Student Member,~IEEE,}
Rafaela~Scaciota\IEEEauthorrefmark{1}\IEEEauthorrefmark{2},~\IEEEmembership{Member,~IEEE,}\\
Sumudu~Samarakoon\IEEEauthorrefmark{1}\IEEEauthorrefmark{2},~\IEEEmembership{Member,~IEEE}, and
Mehdi Bennis\IEEEauthorrefmark{1}~\IEEEmembership{Fellow,~IEEE}
}
\IEEEauthorblockA{
	\small%
	\IEEEauthorrefmark{1}%
	Centre for Wireless Communication, University of Oulu, Finland \\
	 %\\
	\IEEEauthorrefmark{2}%
 Infortech Oulu, University of Oulu, Finland \\
 email: \{kalpana.ranasinghe, madushanka.hewapathiranage, rafaela.scaciotatimoesdasilva, sumudu.samarakoon, mehdi.bennis\}@oulu.fi
}
\vspace{-20pt}
\thanks{
This work was supported by NSF-AKA CRUISE (GA 24304406), VERGE (GA 101096034), 6G-INTENSE (GA 101139266), and Infotech-R2D2. Views and opinions expressed are however those of the author(s) only and do not necessarily reflect those of the European Union. Neither the European Union nor the granting authority can be held responsible for them.} 
}

\maketitle

\begin{abstract}

This work presents a use case of \gls{fl} applied to discovering a maze with LiDAR sensors-equipped robots. % integrated within JetBots.
Goal here is to train classification models to accurately identify the shapes of grid areas within two different square mazes made up with irregular shaped walls.
Due to the use of different shapes for the walls, a classification model trained in one maze that captures its structure does not generalize for the other.
This issue is resolved by adopting \gls{fl} framework between the robots that explore only one maze so that the collective knowledge allows them to operate accurately in the unseen maze. 
%
% and the model consists of 15 possible classes representing different grid shapes.
%
%By leveraging the collective knowledge of two JetBots, trained model is able to generalize across two different maze structures, improving the classification performance for different types of irregular shapes.
%
%Proposed federated learning approach eliminates model overfitting to specific maze, robot characteristics and enables successful classification even when robots swap their mazes. 
%
This illustrates the effectiveness of \gls{fl} in real-world applications in terms of enhancing classification accuracy and robustness in maze discovery tasks.

\end{abstract}

\glsresetall

\section{Introduction}

Robotic navigation in complex environments requires accurate perception and understanding of the surroundings. 
However, when multiple robots operate in different environments independently, their ability to generalize for unseen environments can be limited due to the lack of shared information based on constraints such as communication, storage, privacy, etc. 
\Gls{fl} has emerged as a promising approach to address this challenge by enabling learning agents to collaboratively train machine learning models without sharing raw data \cite{mcmahan2016communication}. 
%, which also leads to less communication overhead. 
In this work, we demonstrate the effectiveness of \gls{fl} in the context of maze discovery and navigation using two autonomous robots in two maze environments with distinct structural builds. 
As oppose to individual learning that fails robots to identify the features of the unseen maze, the \gls{fl}-based approach allows robots to train generalized models that can be accurately used to identify the features in both seen and unseen mazes. 
% With the help of \gls{fl} during the training phase, the robots achieve enhanced generalisation and successfully navigate mazes with distinct characteristics. 
In this view, our results highlight the importance of \gls{fl} in real-world applications in terms of improving the collective knowledge and performance of multiple robots in diverse environments without exchanging the raw data.

\section{Features of Square Grid Mazes}

\begin{figure}[!t]
\centering
\includegraphics[width=1\linewidth]{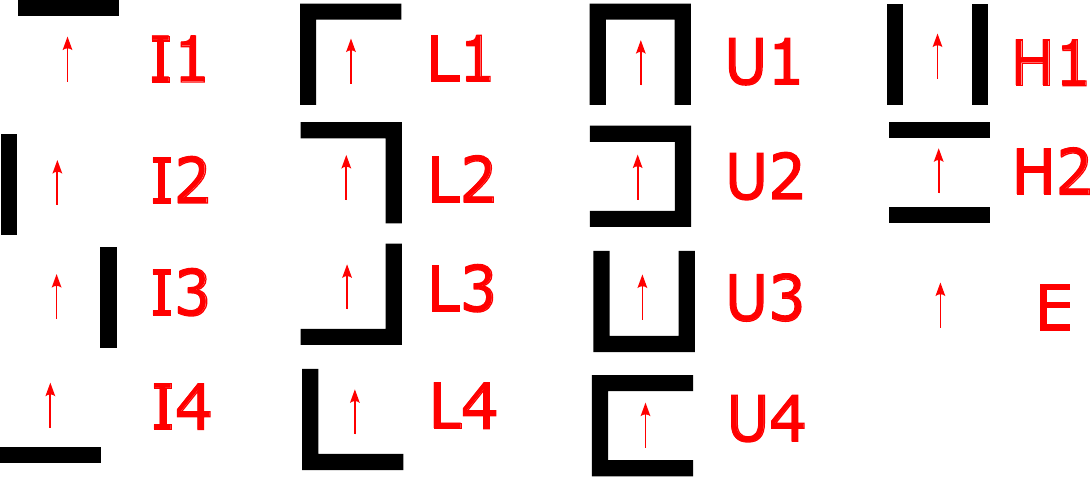}
\caption{Block types present in a maze with respect to the orientation of the observer and their corresponding label.
}
\label{fig:blocks}
\end{figure}

A maze consists of smaller grids, each representing a specific shape within the overall structure. 
In a square-type maze, there are 15 predefined shapes, including open areas as shown in Fig.~\ref{fig:blocks}.
By accurately classifying the block types observed from the middle of a grid using integrated sensory data, robots can navigate and explore the maze.
To achieve this, a robot needs an accurate classification model. 
The nature of the maze, such as the irregularities in wall shapes, can be different from one maze to another. 
Therefore, robots need data from various types of environments to train a robust classification model.

\section{System Architecture \& Hardware}\label{sec:architecture}

\begin{figure}[h]
\centering
\includegraphics[width=\linewidth]{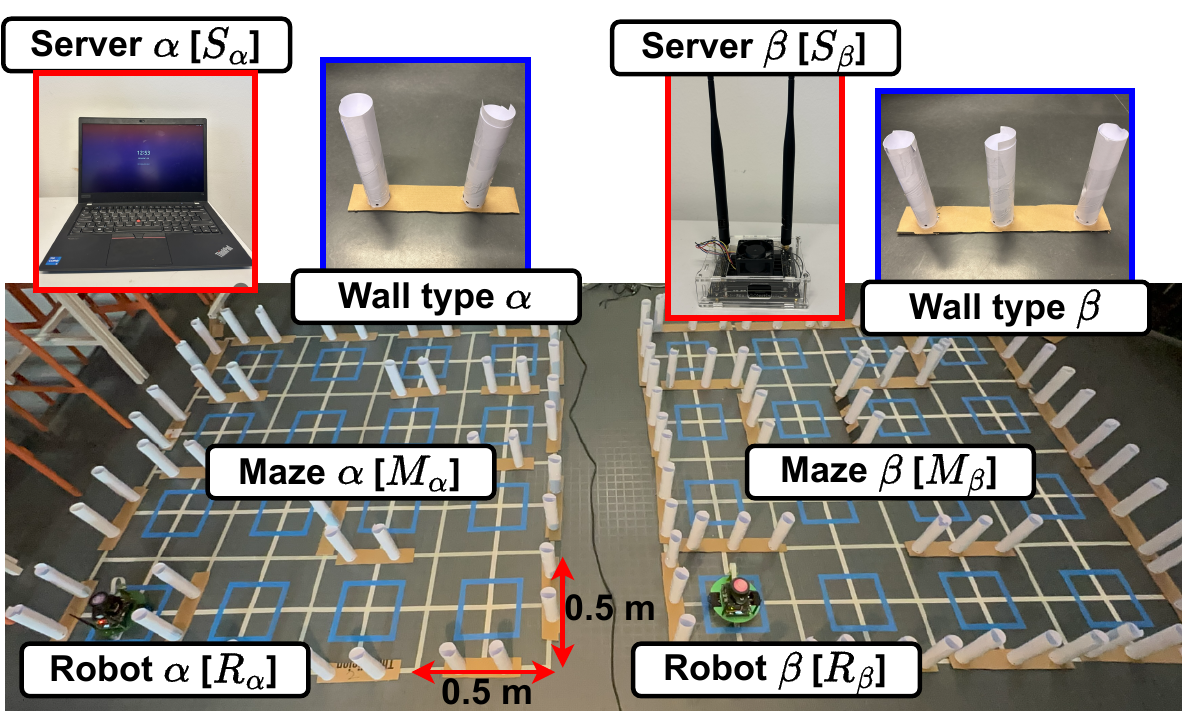}
\caption{Robotic platform of two mazes.}
\label{fig:sysmodel}
\end{figure}

The demo setup consists of two $4\times4$ square grid mazes: Maze $\A$ ($\maze_\A$) and Maze $\A$ ($\maze_\B$), each constructed with unique irregular-shaped walls: type $\A$ and type $\B$, two robots namely $\robot_\A$ and $\robot_\B$ that use LiDAR sensors to perceive the environment, and two slave edge servers referred to as $\server_\A$ and $\server_\B$ as shown in the Fig.~\ref{fig:sysmodel}. 
% \sps{The dimensions of one grid in the maze is 0.5\,m*0.5\,m}{This can be shown in the figure itself so no need to add it in the text.} 
%
Wall type $\A$ consists of a base with two cylindrical shapes attached to it, while wall type $\B$ is characterized by a base with three cylindrical shapes.

Operating as \gls{ros} masters, $\robot_\A$ and $\robot_\B$ are assigned the tasks of navigation, data collection, and maze discovery within their respective mazes $\maze_\A$ and $\maze_\B$.
%
% JetBot alpha and JetBot beta are assigned to discover the $M_A$ and $M_B$, respectively.
%
Therein, the robots are positioned at the starting point $(0,0)$ in each maze. 
%They employ LiDAR sensors to gather information about the surrounding environment for different types of walls.
%
As for mobile robots, we use open-source robots based on NVIDIA Jetson Nano known as "JetBot \gls{ros} \gls{ai} kit" \cite{jetbot}.
Each robot is equipped with a $360^\circ$ laser ranging LiDAR that is used to observe the surrounding from the middle of the grid, and a $8$\,MP $160^\circ$ field of view camera.
To achieve precise navigation between grid centers, robots use a line-following technique.
Utilizing cameras mounted on the robots, they track white lines marked on the floor, ensuring accurate movement along the paths. 
Additionally, blue lines are marked in front of each grid center, enabling the robots to stop precisely at the center. % as later discussed under~\ref{sec:navigation}. 
%
% By aligning with these blue lines, the robots achieve accurate positioning within the maze, facilitating efficient exploration.
%
The LiDAR model is RPLIDAR A1, which has a scanning frequency of $5.5$\,Hz, a ranging distance of $0.15\sim12$\,m with an accuracy of $1\%$ for distances less than $3$\,m, and $1147$ sample points per one sweep.
To obtain accurate odometry information, the robot utilizes two high-power encoded motors, an IMU sensor, and an odometer. 
These sensors work together to provide reliable odometry data, which is then represented in the \gls{ros} framework.

% As mobile robots, two JetBots, JetBot alpha and JetBot beta, are deployed as ROS masters, acting as autonomous agents responsible for navigation and data collection within their respective mazes. Positioned at the starting point (0,0) in each maze, the JetBots use their LiDAR sensors, to gather information about their surroundings. 
%
% JetBot alpha and JetBot beta expected to discover the Maze A and Maze B respectively. 
% By capturing LiDAR readings from the middle of each grid, the JetBots can accurately identify the block types present.
%
The slave servers $\server_\A$ and $\server_\B$ are used to facilitate visualization and monitoring of the maze discovery process.
A Lenovo Thinkpad with \gls{ros} is used as $\server_\A$ and a Jetson Nano with \gls{ros} is used as $\server_\B$.
%
% These servers receive data from the robots and generate graphical user interfaces (GUIs) that visualize the maze layout, robot positions and orientations, and the identified block types in real-time. 
%The slave servers provide real-time visualisation and analysis, enabling us to observe the maze discovery progress and make informed decisions based on the collected data in the inference phase.
%
The slave servers and the robots use \gls{ros} topics for seamless and efficient communication. 
The robots publish their sensor readings and navigation commands, while the slave servers subscribe to these topics to receive the data. 
These data includes position and direction of the robot, identified block type, and flag messages for synchronization between \gls{ros} nodes, enabling the slave servers to visualize the robots' progress, block identifications, and navigation decisions over a graphical user interface.

\section{System Implementation}

This demo is composed of several components including the navigation, data collection, training of classification models, and inference and maze discovery as discussed next.

\subsection{Navigation between adjacent grids}
\label{sec:navigation}
%TODO: Madushanka is updating this section
A vision-based line-following system is implemented on the robots to facilitate successful navigation between the grids using the camera serial interface camera module.
The camera, along with the Jetson Nano, supports image processing at resolutions of $480\times640$ pixels, with a maximum frame rate of $30$ frames per second. 
The navigation process consists of two repeated steps: i) the robot follows the white line, and ii) the robot stops at the center of each grid by detecting the blue line.

We choose a \gls{roi} from the bottom portion of the raw camera image to prioritise the detection of nearby points during line following.
The \gls{roi} is then converted from RGB to HSV color space, due to the effectiveness of filtering out white lines based on the luminosity. 
%
%with subsequent adjustments made to the H, S, and V values until the white line is effectively filtered. 
%Because HSV color spaces separate image luminance from color information.
%
Subsequently, the contour with the largest area is chosen to accurately identify the white line. 
The coordinates of the center point of this contour is extracted and used to determine the error, which represents the displacement of the path's center from the center of the \gls{roi}~\cite{openCV_contour}. 
This error value is used as an input for a \gls{pid} controller, ensuring precise control commands and smooth navigation~\cite{pid_controller}. 
The coefficients of the proportional, derivative, and integral terms in the \gls{pid} controller are set to $3$, $1.8$, and $0$, respectively.

Detection of the blue line is used to stop the robot precisely at the center of each grid. 
The blue line detection process is similar to the white line detection, with exceptions of the use of a different \gls{roi} and the HSV adjustments to filter out the blue lines. 
%from the raw image was chosen.
%
%Subsequently, the \gls{roi} was converted to HSV color space, and the H, S, and V values were adjusted to effectively filter the blue line. The contour with the largest area was then selected to precisely identify the blue line. 
The robot stops once the detected area of the blue line exceeded a predetermined threshold value, thereby achieving precise positioning at the center of the grid.
%
% These methods and techniques demonstrate the successful implementation of the vision-based line-following system on the Jetbot, enabling accurate navigation and controlled stopping procedures.

Further, we use the odometer frame information from the \texttt{/odom} rostopic to make precise $90\degree$ and $180\degree$ rotations while navigating by feeding the error obtained from the odometry information and target angle to a \gls{pid} controller~\cite{pid_controller}.

\subsection{Data Collection}
\label{sec:data_collect}

During the data collection phase, we feed a predefined path and the map information as an array to the robot assigned for the maze. 
While following the path, at each grid, the robot performs four $90\degree$ rotations and collects $200$ LiDAR sweeps at each position by reading the \texttt{LaserSacn} message from \texttt{/scan} rostopic.
These LiDAR readings are then saved into a numpy array with the corresponding label as shown in the Fig.~\ref{fig:blocks}.
% \sps{correct label}{What is the correct label? Please introduce the convention. You can use Fig. 1 to show the sphape and define the label easily.}. 
% \sps{LiDAR in the mobile robot is RPLIDAR A1 which has a scanning frequency of 5.5 Hz, ranging distance of 0.15~12.00m, ranging accuracy of 1\% for distances less than 3 m and it has 1147 sample points per one sweep.}{Should not this be in HW discussions?}
%
To enhance the diversity of the dataset within a short time frame, 
some noises to the movements are manually fed to the robot via a remote controller 
while it collects data from a single block type. 
%\kpr{we feed manual small movements}{Then how the predefined path is utilized? Movements are small and within one grid. In the end make robot to be in the initial position using remote before it starts rotating to collect data for the next block} to the robot with the help of remote controller within the grid while the robot collects data for a single block type. 
%
These variations aid to generate a comprehensive local dataset capturing a wide range of environmental configurations and scenarios, enhancing the overall robustness of the trained model.
%
% During the data collection phase, JetBot Alpha is responsible for collecting data from Maze A, while JetBot Beta collects data from Maze B. Each JetBot creates its own dataset to train local classification models specific to their respective mazes.

%

\subsection{Training the classification model}

% \sps{NOTE}{Say that we use two modes: local training and FL. This is useful for the discusJetBotsion under inference.}
We use two training modes to train the classification model: \emph{local training} and \emph{\gls{fl}}.
During the local training mode, both robots train their classification models locally using the datasets obtained from respective mazes with supervised learning. 
The classification models are based on a feed-forward neural network architecture, consisting of a $1147\times 1$ input reflecting LiDAR reading, one hidden layer with a size of 256 neurons, and a $15\times 1$ output layer corresponding one-hot encoded label with ReLu activation.
To prevent over-fitting and improve generalization, a learning rate of $0.001$ and the L2 regularization  with a weight decay of $0.001$ are applied in the training.
After the training process, both models achieve a classification accuracy of $99\%$ on their local testing data.
%

\if0
This high accuracy demonstrates the effectiveness of the chosen model architecture and the quality of the collected datasets in capturing the distinctive characteristics of the mazes.
%
\sps{--}{Is not the following a part of inference?}
\kpr{Kind of}{But here added these before going into FL part with a flow. In the inference/maze discovery section mainly focusing on maze discovery rather than numerical values}
However, testing the locally trained models on the test data from the other maze reveals a significant drop in accuracy.
For $\robot_\A$, the test accuracy is about $48\%$ on the data from $\maze_\B$ while $\robot_\B$ achieves merely 29\% accuracy on $\maze_\A$'s test data. 
These results highlight the need for collaborative learning to overcome the limitations of individual models.
\fi

% \sps{}{Introduce how FL is implemented here. What acts as the server? What are the FL-related parameters?}
During the \gls{fl} mode, $\robot_\A$ and $\robot_\B$ act as clients, while $\server_\B$ acts as the parameter server, and \emph{FedAvg} algorithm~\cite{mcmahan2016communication} is used for the training.
In the \gls{fl} mode, we used the same model architecture and learning rate as in the local training mode, with a minibatch size of $16$, two local iterations prior to model averaging, and $15$ global iterations.
Under \gls{fl}, the global model used in both robots achieve a test accuracy of about $99\%$.

%Accuracy over the combined test dataset of Maze A and Maze B significantly improves to 99\% in the \gls{fl} mode. This demonstrates the effectiveness of \gls{fl} in achieving a highly accurate classification model that can generalize across different environments. 
%
% A summary of the \gls{fl} process is depicted in Fig.~\ref{fig:fl_stats}

\subsection{Maze discovery}

\begin{figure}[!t]
\centering
\includegraphics[width=0.8\linewidth]{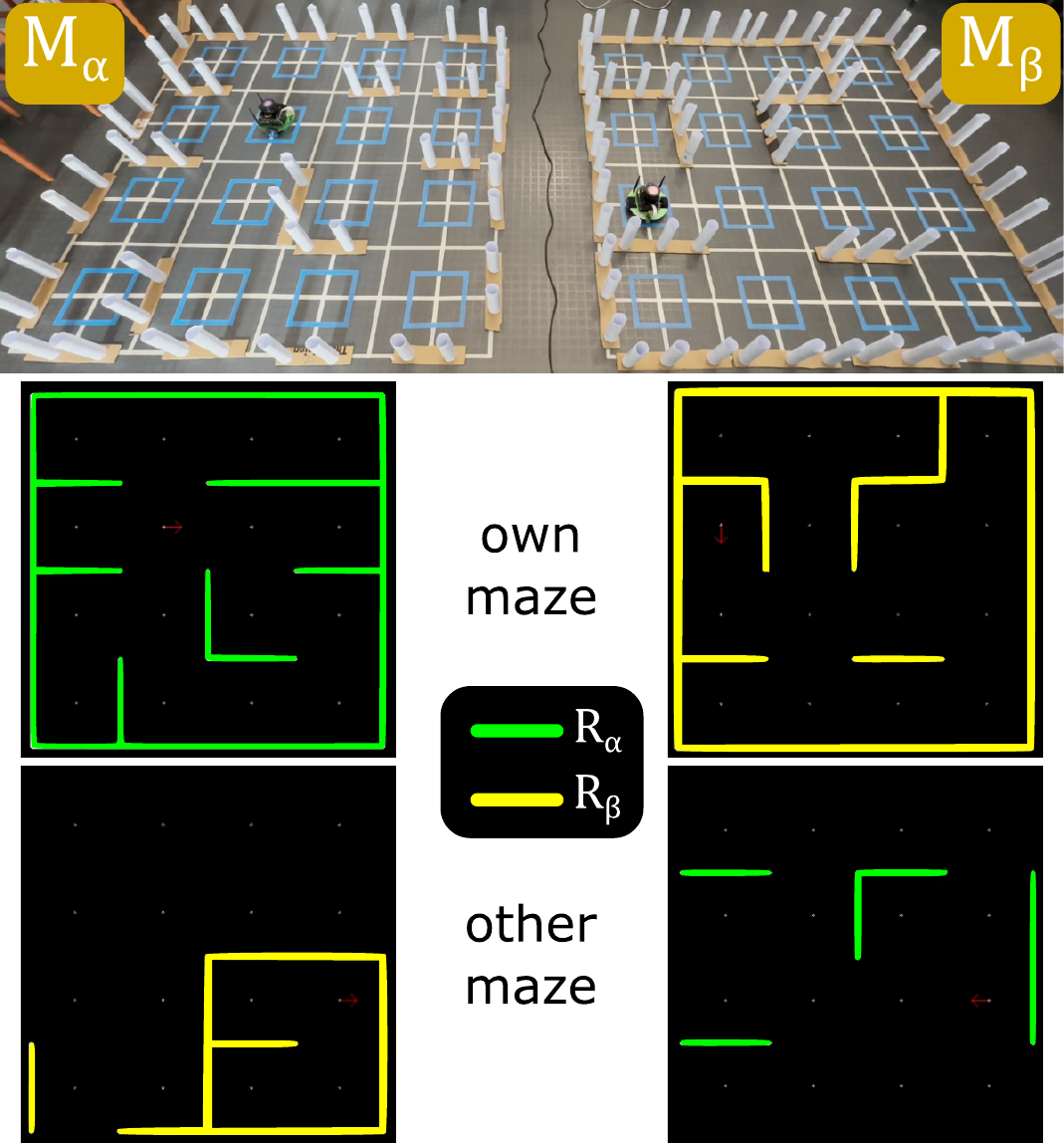}
\caption{Discovering mazes (top) using locally trained models: inference output of the own maze (middle) vs the other maze (bottom). }
%\caption{Maze discovery with locally trained models}
\label{fig:local}
\end{figure}

The inference of the classification models obtained by local training and \gls{fl} modes is used within the discovery phase as two different scenarios.
% \sps{[mode A] and [mode B]}{This revision is wrong. Define the modes and explain the training process for these two modes.}.
Therein, each robot self navigates within the mazes by taking different actions with the objective of discovering the maze layout. 
The action space includes four actions correspond to moving forward after a clockwise rotation of $0\degree$, $90\degree$, $-90\degree$, or $180\degree$.
%, rotating $90\degree$ clockwise and moving forward, rotating $180\degree$ and moving forward, and rotating $90^\circ$ counterclockwise and moving forward. 
To ensure efficient maze exploration, a prioritized clockwise turn strategy is employed to avoid unnecessary looping. 
Once the maze is discovered, the robot automatically stops at the adjacent grid. 
The navigation techniques described in Section \ref{sec:navigation} are utilized to navigate between grid centers.

Fig.~\ref{fig:local} illustrates the maze discovery using the locally trained models.
It can be noted that $\robot_\A$ and $\robot_\B$ can correctly infer their corresponding mazes $\maze_\A$ and $\maze_\B$, respectively validating the test accuracy obtained during the training phase. 
However, once the robots are swapped, they no longer able to correctly identify the unseen wall types, in which, maze discovery is inaccurate as shown in the bottom row of Fig.~\ref{fig:local}.
With further experiments, it is observed that $\robot_\A$ exhibits about $48\%$ of classification accuracy on the data from $\maze_\B$ while $\robot_\B$ achieves merely $29\%$ accuracy on $\maze_\A$'s data. 
%These results highlight the need for collaborative learning to overcome the limitations of individual models.
This indicates that the local training does not generalize for unseen environments.

%\begin{figure}[!h]
%\centering
%\includegraphics[width=\linewidth]{Figures/local_swapped.pdf}
%\caption{Maze discovery with locally trained models swapped
%System Architecture. 
%}
%\label{fig:localswapped}
%\end{figure}

%However, when the robots are swapped, as depicted in Fig.~\ref{fig:localswapped}, they fail to accurately discover the respective mazes.
%

\begin{figure}[!t]
\centering
\includegraphics[width=0.8\linewidth]{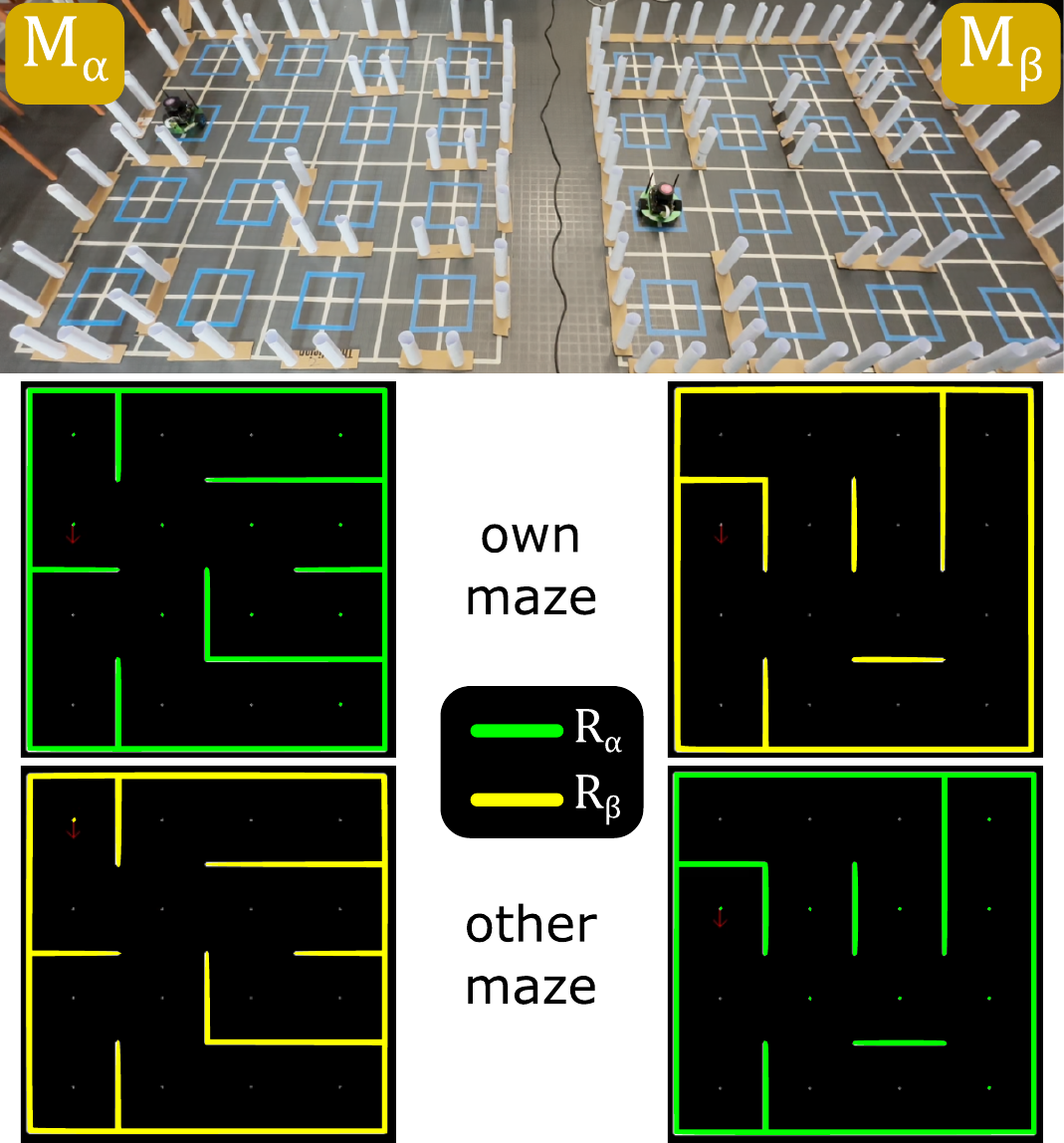}
\caption{Discovering mazes (top) using FL models: inference output of the own maze (middle) vs the other maze (bottom). }
\label{fig:fl}
\end{figure}

The inference using the \gls{fl} models are illustrated in Fig. \ref{fig:fl}.
In contrast to local training alone, with \gls{fl}, the robots are capable of sharing their local knowledge with one another, in which, the models are generalized for unseen wall types.
This is exhibited by the accurate maze discovery of both mazes by the both robots as shown in the inferred outputs (middle and bottom) of Fig. \ref{fig:fl}.
Our experiments show that the \gls{fl}-based model achieves about $99\%$ classification accuracy over the data from both $\maze_\A$ and $\maze_\B$.

\subsection{Demo, Resources, and Future Developments}

The software related to robots and servers is available at \url{https://github.com/ICONgroupCWC/MapDiscoveryDemo}.
This demo in action can be seen from \url{https://youtu.be/K2M8MCLn1po?si=TzIeUhHuSVCVeZRl}
The future developments are focused on exploiting the topological signatures of LiDAR point-cloud data to train models over the symmetries, deformations, and irregularities in the raw observations with improved communication and computation efficiency.

% --------------- WORKS CITED (10pt FONT) ---------------------
\bibliographystyle{IEEEtran}
\bibliography{reference}

\begin{thebibliography}{1}
\providecommand{\url}[1]{#1}
\csname url@samestyle\endcsname
\providecommand{\newblock}{\relax}
\providecommand{\bibinfo}[2]{#2}
\providecommand{\BIBentrySTDinterwordspacing}{\spaceskip=0pt\relax}
\providecommand{\BIBentryALTinterwordstretchfactor}{4}
\providecommand{\BIBentryALTinterwordspacing}{\spaceskip=\fontdimen2\font plus
\BIBentryALTinterwordstretchfactor\fontdimen3\font minus
  \fontdimen4\font\relax}
\providecommand{\BIBforeignlanguage}[2]{{%
\expandafter\ifx\csname l@#1\endcsname\relax
\typeout{** WARNING: IEEEtran.bst: No hyphenation pattern has been}%
\typeout{** loaded for the language `#1'. Using the pattern for}%
\typeout{** the default language instead.}%
\else
\language=\csname l@#1\endcsname
\fi
#2}}
\providecommand{\BIBdecl}{\relax}
\BIBdecl

\bibitem{mcmahan2016communication}
H.~B. McMahan \emph{et~al.}, ``Communication-efficient learning of deep
  networks from decentralized data. arxiv,'' \emph{arXiv preprint
  arXiv:1602.05629}, 2016.

\bibitem{jetbot}
Waveshare, ``Jetbot ros ai robot,'' Available at
  \url{https://www.waveshare.com/jetbot-ros-ai-kit.htm} (2023/07/10).

\bibitem{openCV_contour}
OpenCV, ``Contours in opencv,'' Available at
  \url{https://docs.opencv.org/4.0.0/dd/d49/tutorial_py_contour_features.html}
  (2023/07/12).

\bibitem{pid_controller}
S.~Bennett, ``Development of the \uppercase{PID} controller,'' \emph{IEEE
  Control Systems Magazine}, vol.~13, no.~6, pp. 58--62, 1993.

\end{thebibliography}

\end{document}